\documentclass[11pt]{article}

\usepackage{authblk}       
\usepackage{graphicx}      
\usepackage{amsmath, amssymb}
\usepackage{hyperref}
\usepackage{natbib}        
\usepackage{listings}      
\usepackage{geometry}   
\usepackage{float}
\title{StructuralDecompose: A Modular Framework for Robust Time Series Decomposition in R}

\author[1]{Allen Daniel Sunny}
\affil[1]{University of Maryland, College Park \\
\texttt{allens@umd.edu}}

\date{\today}

\begin{document}

\maketitle

\begin{abstract}
We present \texttt{StructuralDecompose}, an R package for modular and interpretable time series decomposition. 
Unlike existing approaches that treat decomposition as a monolithic process, 
\texttt{StructuralDecompose} separates the analysis into distinct components: changepoint detection, 
anomaly detection, smoothing, and decomposition. This design provides flexibility and robustness, 
allowing users to tailor methods to specific time series characteristics. 
We demonstrate the package on simulated and real-world datasets, benchmark its performance against state-of-the-art tools such as Rbeast and autostsm, 
and discuss its role in interpretable machine learning workflows.
\end{abstract}

\section{Introduction}
Understanding time series data is central to domains such as economics, finance, health, and policy. 
Beyond predictive accuracy, many applications increasingly require \textit{explainable and interpretable} methods 
that can reveal underlying temporal structures in a transparent manner. 
Interpretable decompositions help practitioners understand long-term trends, identify seasonal patterns, 
and detect anomalies or regime changes that drive real-world outcomes. 
For decision-making in high-stakes domains, such as economic forecasting or public health surveillance, 
black-box models without interpretable structure can be difficult to trust or validate.  

Classical approaches such as STL \cite{wen_robuststl_2018} and state-space models \cite{zhang_effectively_2023} 
provide interpretable decompositions into trend, seasonal, and residual components. 
However, these methods often assume stable underlying dynamics and are sensitive to anomalies or structural breaks. 
Recent methods such as Rbeast \cite{zhao_detecting_2019} and autostsm  \cite{hubbard_autostsm_2021} 
introduce scalable or Bayesian approaches but largely treat decomposition as a single, monolithic process. 
This limits the ability of analysts to tailor methods for specific explanatory goals.  

\texttt{StructuralDecompose} introduces a modular framework designed with interpretability in mind. 
By decoupling decomposition into distinct components---changepoint detection, anomaly handling, smoothing, 
and seasonal/residual decomposition---the package provides transparency at each stage of analysis. 
Users can explicitly observe where breaks occur, which anomalies are flagged, how smoothing affects trend estimates, 
and how seasonal structures are derived. This level of control supports reproducibility and explainability, 
bridging the gap between traditional statistical decomposition and modern demands for interpretable machine learning.

\section{Related Work}
Time series decomposition has been approached from several methodological traditions, 
each with its own strengths and limitations. We highlight three major families: Bayesian methods, 
state-space models, and classical statistical decomposition. 

\subsection{Other Time Series Methods}
Time series decomposition has traditionally been approached using a variety of methods, 
including Bayesian models, state-space formulations, and classical statistical techniques.  Bayesian approaches, such as \texttt{Rbeast} \cite{zhao_detecting_2019}, provide probabilistic 
decompositions with explicit uncertainty estimates, but are computationally intensive 
and often less transparent for practitioners. State-space models \cite{zhang_effectively_2023}, 
and their automated extensions like \texttt{autostsm} \cite{hubbard_autostsm_2021}, represent 
unobserved components through latent states estimated via the Kalman filter, making them 
flexible for forecasting but harder to interpret. Classical approaches such as STL 
decomposition \cite{wen_robuststl_2018} remain popular for their simplicity and transparency, 
but are less robust to anomalies and structural breaks.  

While each of these methods has strengths, they generally treat decomposition as a 
monolithic process and provide limited modularity. In contrast, \texttt{StructuralDecompose} 
emphasizes interpretability by exposing changepoint detection, anomaly handling, smoothing, 
and decomposition as distinct, customizable steps.

\subsection{Changepoint Detection}
Changepoint detection has been an active area of research in time series analysis, 
with methods differing in their statistical assumptions, computational efficiency, 
and robustness. Early approaches relied on cumulative sum (CUSUM) statistics and 
likelihood ratio tests to identify mean or variance shifts in Gaussian series. 
Recursive partitioning methods such as binary segmentation \cite{fryzlewicz_wild_2014}
remain widely used for their simplicity, although they may overestimate the 
number of changepoints.  

More recent advances include the Pruned Exact Linear Time (PELT) algorithm 
\cite{killick_optimal_2012}, which uses dynamic programming with pruning rules to achieve 
near-linear scalability while providing exact solutions under penalized cost 
functions. Segment neighborhood methods offer exact inference but are limited to 
shorter series due to their quadratic complexity. Nonparametric approaches, such 
as energy statistics and kernel methods \cite{garreau_consistent_2017}, extend changepoint 
detection beyond mean and variance shifts to capture distributional changes.  

In practice, software implementations such as the R packages \texttt{changepoint} 
and \texttt{strucchange} provide accessible interfaces to these algorithms. 
\texttt{StructuralDecompose} integrates with these libraries, allowing users to 
select the most appropriate changepoint method for their application. Breakpoints 
are then used to segment the series, enabling anomaly detection and smoothing 
to adapt to local structure.

\subsection{Anomaly Detection}
Anomaly detection, also referred to as outlier detection, is another well-studied 
problem in time series analysis. Early approaches focused on statistical thresholding, 
such as z-score methods, where points are flagged if they deviate from the mean by 
a multiple of the standard deviation. Robust alternatives based on the Median Absolute 
Deviation (MAD) \citep{hochenbaum_automatic_2017} offer greater resistance to extreme values.  

Moving-window techniques, such as rolling averages or robust regression residuals, 
extend detection to nonstationary series by comparing local statistics over time. 
Energy-based and density-based methods, including isolation forests and one-class SVMs 
\cite{chalapathy_anomaly_2019}, have been adapted to time series to capture nonlinear or 
distributional anomalies. More recent developments apply deep learning, such as 
autoencoders and recurrent neural networks, to learn representations of “normal” 
patterns and flag deviations \cite{radford_network_2018}.  

Within the R ecosystem, packages such as \texttt{anomalize} provide tidyverse-friendly 
implementations of statistical and regression-based methods, while \texttt{tsoutliers} 
integrates anomaly detection with ARIMA modeling. \texttt{StructuralDecompose} adopts 
a modular approach, supporting z-score, MAD, and window-based methods natively, while 
allowing users to plug in external anomaly detection routines before smoothing and 
decomposition. This ensures that anomalies do not distort trend or seasonal estimates, 
and their handling remains transparent to the analyst.

\subsection{Trend Smoothing}
Smoothing is a central task in time series analysis, used to estimate long-term trends 
while filtering out short-term fluctuations. Classical approaches include moving averages 
and exponential smoothing \cite{abrami_time_2017}, which are computationally simple and widely 
used in forecasting. LOESS (locally estimated scatterplot smoothing) \cite{smolik_comparative_2016}
introduced nonparametric local regression, offering flexibility and interpretability by 
fitting low-degree polynomials to localized subsets of the data. Splines and penalized 
regression approaches provide another flexible alternative, balancing smoothness and fit 
through regularization \cite{lenz_adaptive_2022}.  

State-space models offer a probabilistic formulation of trend smoothing, where the latent 
trend evolves according to a stochastic process and is estimated via the Kalman filter and 
smoother \citep{pei_elementary_2019}. While this provides uncertainty quantification and a principled 
forecasting framework, it is often less transparent to practitioners, as the latent state 
representation and variance parameters can obscure the effect of smoothing.  

In practice, R implementations of smoothing include \texttt{stats::filter()} for moving 
averages, \texttt{loess()} for local regression, and \texttt{smooth.spline()} for spline 
fitting. \texttt{StructuralDecompose} emphasizes LOESS as a default due to its 
interpretability and direct connection to the well-known STL decomposition, while also 
supporting splines and moving averages. By making the smoother explicit and user-selectable, 
the package avoids the opacity of latent-state approaches and provides transparent control 
over trend estimation.

\subsection{Decomposition}
Decomposition aims to represent a time series as the sum of interpretable components, 
typically expressed as
\[
y_t = T_t + S_t + R_t,
\]
where $T_t$ is the long-term trend, $S_t$ the seasonal component, and $R_t$ the residual. 
Classical decomposition methods in R, such as \texttt{decompose()} and \texttt{stl()}, 
separate these components using moving averages or LOESS smoothing \citep{adams_moving_2023}. 
STL (Seasonal-Trend decomposition using LOESS) remains particularly popular for its 
flexibility in handling nonlinear seasonality and its intuitive interpretation.  

Extensions of decomposition include multiplicative models ($y_t = T_t \times S_t \times R_t$), 
useful when seasonal variation grows with trend, and robust decomposition methods that reduce 
the influence of outliers \cite{zhang_detection_2025}. State-space formulations provide another view, 
where trend and seasonality are modeled as latent states estimated via the Kalman filter 
\cite{pei_elementary_2019}, though these approaches are often more complex and less transparent for 
exploratory analysis.  

In the R ecosystem, decomposition is available in base functions (\texttt{decompose}, 
\texttt{stl}), as well as packages such as \texttt{forecast} and \texttt{prophet}, which 
emphasize forecasting. However, these methods typically treat decomposition as a single 
operation. By contrast, \texttt{StructuralDecompose} makes decomposition the final stage 
of a modular pipeline, applied only after changepoints, anomalies, and smoothing are 
explicitly handled. This ensures that seasonal and residual components are derived from a 
cleaner, more interpretable signal.

\section{Methods}

\texttt{StructuralDecompose} is implemented as a modular R framework that processes a time series in four explicit stages: \textbf{changepoint detection}, \textbf{anomaly detection}, \textbf{trend smoothing}, and \textbf{final decomposition}. Each stage is designed to be independent, user-selectable, and transparent, enabling analysts to control how the series is prepared and interpreted.

\subsection{Changepoint Detection}
By default, the package leverages the \texttt{strucchange} library to detect structural breaks. It also integrates with \texttt{changepoints} and \texttt{segmented}, allowing users to apply methods such as binary segmentation, PELT, or CUSUM. Identified changepoints are stored as indices and used to partition the series before subsequent processing. Results can be inspected using the \texttt{BreakPoints} function, which returns the detected break indices.

\begin{figure}[H]
    \centering
    \includegraphics[width=0.5\linewidth]{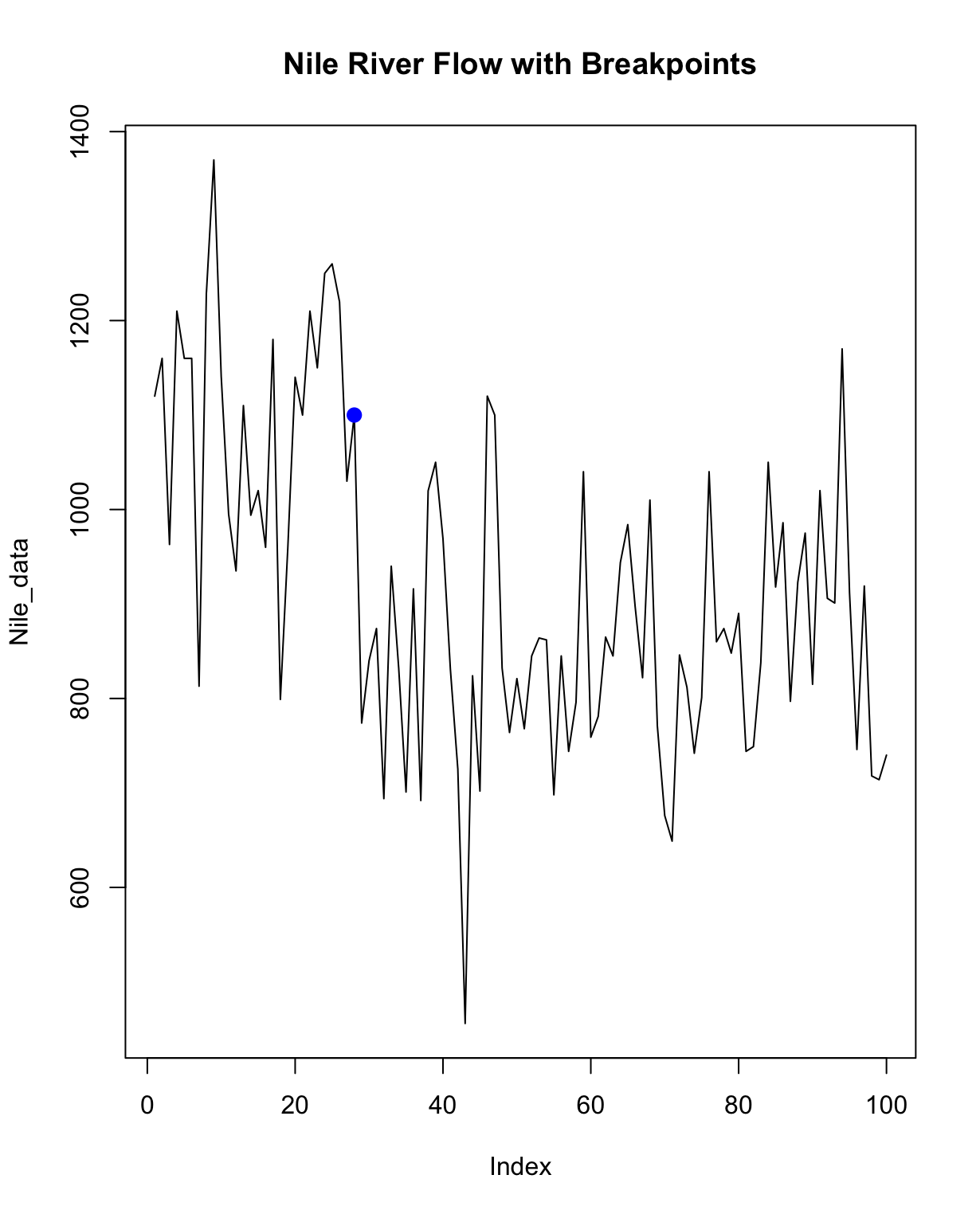}
    \label{fig:placeholder}
\end{figure}

\subsection{Anomaly Detection}
Within each segment, anomalies are identified and optionally removed to prevent distortion of trend and seasonal estimates. The default approach is a moving-window rolling median filter. Additional built-in methods include z-score filtering, median absolute deviation (MAD), and rolling-window statistics. Detected anomalies can be excluded or replaced according to user-defined rules, ensuring robustness in downstream analysis.
\begin{figure}[H]
    \centering
    \includegraphics[width=0.5\linewidth]{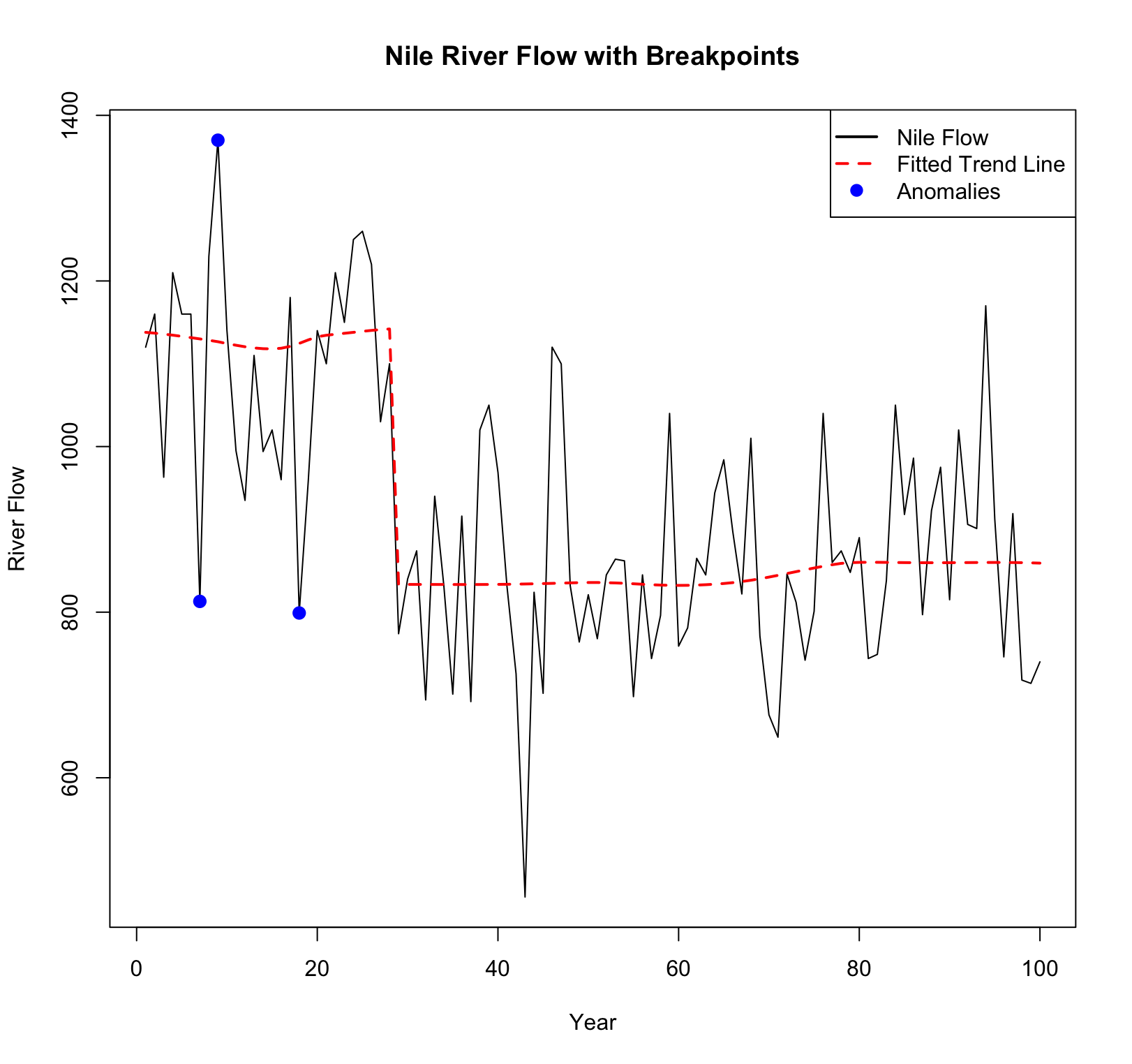}
    \label{fig:Anomaly}
\end{figure}

\subsection{Trend Smoothing}
After segmentation and anomaly handling, smoothing is applied to estimate the trend component on a per-segment basis. Supported methods include \texttt{lowess} (default), moving averages, and splines. Lowess is emphasized for its interpretability, while alternative smoothers provide additional flexibility.
\begin{figure}[H]
    \centering
    \includegraphics[width=0.5\linewidth]{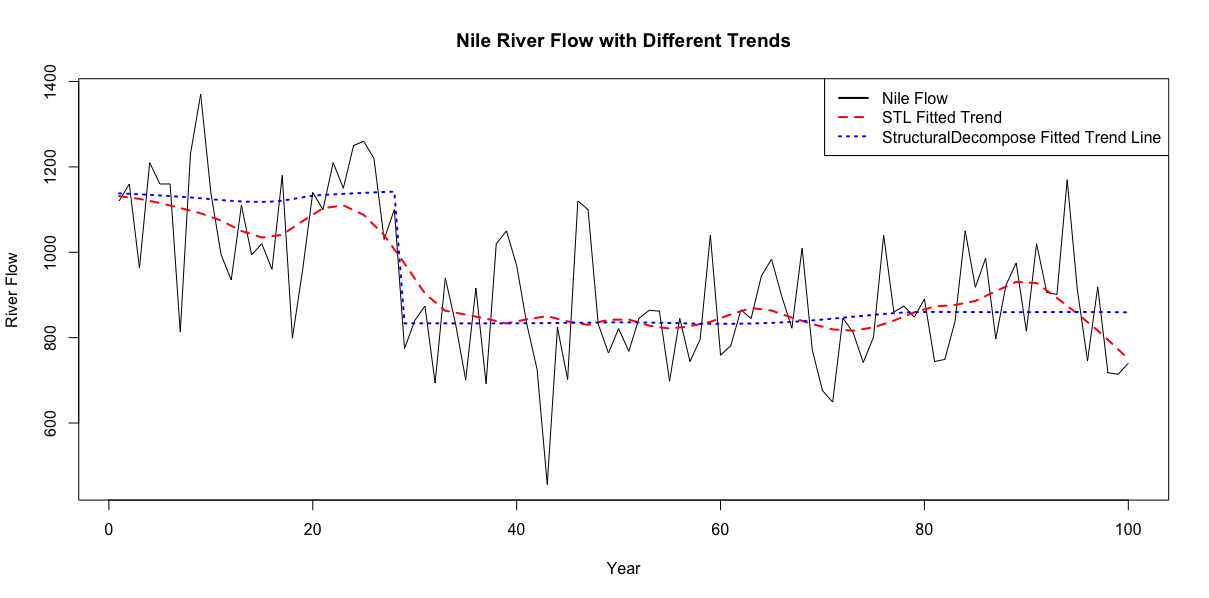}
    \label{fig:Trends}
\end{figure}

\subsection{Decomposition}
Once the trend is estimated, the series is decomposed into trend, seasonal, and residual components:
\[
y_t = T_t + S_t + R_t,
\]
where $T_t$ denotes the trend, $S_t$ the seasonal component, and $R_t$ the residual. The seasonal component is estimated using an STL-style extraction, and the residual is defined as the remainder after removing both trend and seasonality. By treating decomposition as the final stage, the package ensures that trend and seasonality are derived from a cleaned, well-structured signal.

The resulting object contains the detected changepoints, flagged anomalies, trend, seasonal, 
and residual components, along with plotting and summary methods. The modular design allows 
analysts to inspect and adjust each stage, making the process more transparent than 
monolithic decomposition functions.
\begin{figure}[H]
    \centering
    \includegraphics[width=0.5\linewidth]{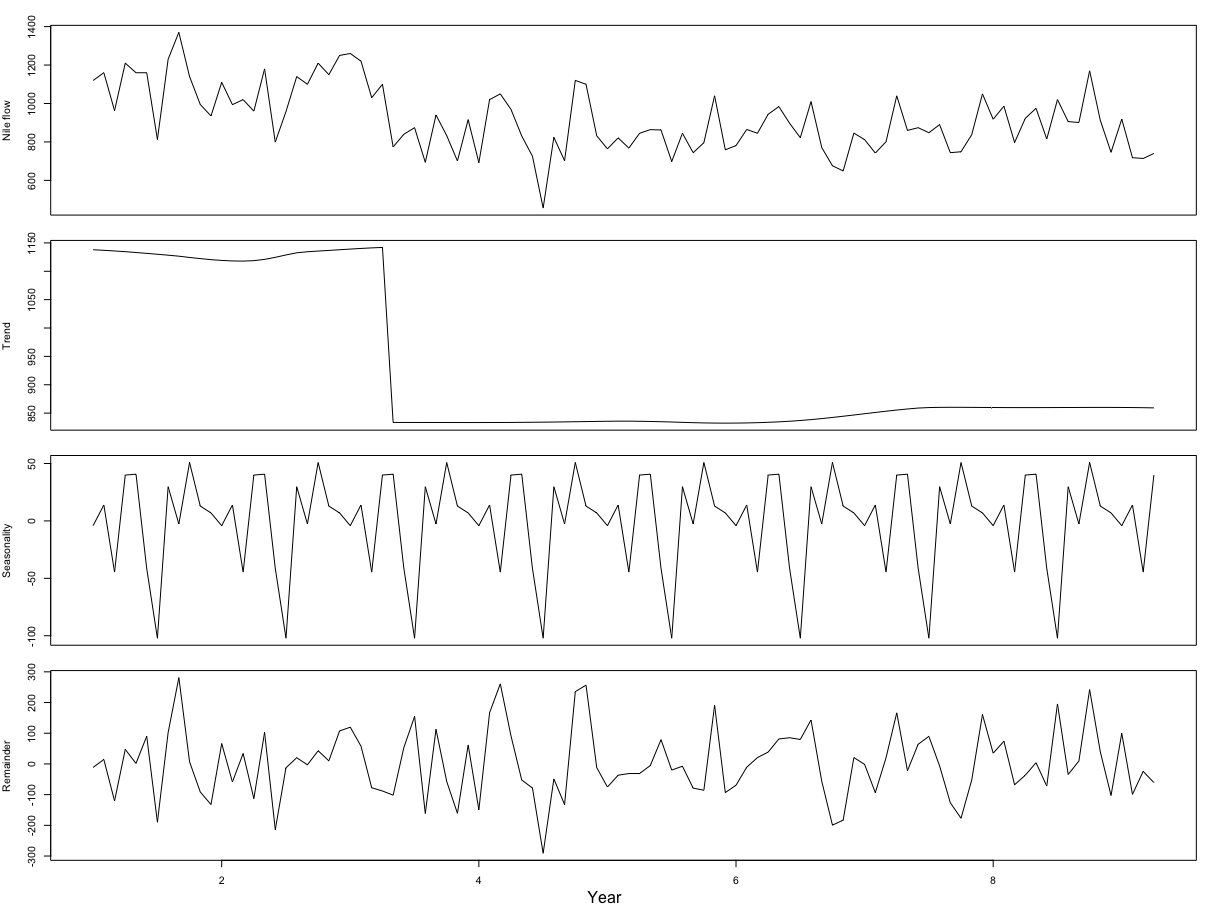}
    \label{fig:Decomposition}
\end{figure}

\section{Software Overview}
The package provides:
\begin{itemize}
    \item \texttt{decompose\_structural()}: main entry point
    \item \texttt{plot\_components()}: visualization helper
    \item \texttt{compare\_methods()}: benchmarking utility
\end{itemize}

Example usage:
\begin{lstlisting}[language=R]
library(StructuralDecompose)

result <- decompose_structural(ts_data, 
                               smoother = "loess",
                               breakpoint = "cusum",
                               anomaly = "zscore")

plot_components(result)
\end{lstlisting}

\section{Software Availability}
Available at: \url{https://github.com/allen-1242/StructuralDecompose}  
License: MIT  

\section{Conclusion}
By decoupling changepoint detection, anomaly handling, smoothing, and decomposition, 
\texttt{StructuralDecompose} provides a flexible and interpretable framework 
for analyzing complex time series.

\bibliographystyle{plainnat}
\bibliography{references}

\end{document}